\documentclass[10pt,journal,compsoc]{IEEEtran}
\usepackage[utf8]{inputenc}
\usepackage{hyperref}
\usepackage{multirow}
\usepackage{amsmath}
\usepackage{amsfonts}
\usepackage{graphicx}
\usepackage{subcaption}
\usepackage{color}	

\ifCLASSOPTIONcompsoc
  \usepackage[nocompress]{cite}
\else
  \usepackage{cite}
\fi

\ifCLASSINFOpdf
\else
\fi

\begin{document}
%
\title{Reinforcement Learning in Conflicting Environments for Autonomous Vehicles}
%
%
%
%

\author{Dominik~Meyer, 
        Johannes~Feldmaier
        and~Hao~Shen
\IEEEcompsocitemizethanks{\IEEEcompsocthanksitem The authors are with the
    Chair for Data Processing, 
    Department of Electrical and Computer Engineering, 
    Technische Universität München.\protect\\
E-mail: $<$firstname.lastname$>$@tum.de.}
\thanks{Manuscript submitted June 13, 2016.}}

%
%

\markboth{Robotics in the 21$^{st}$ century: Challenges and Promises}%
{Meyer \MakeLowercase{\textit{et al.}}: Reinforcement Learning in Conflicting 
Environments for Autonomous Vehicles.}
%



\IEEEtitleabstractindextext{%
\begin{abstract}
In this work, we investigate the application of Reinforcement Learning to
two well known decision dilemmas, namely Newcomb's Problem and
Prisoner's Dilemma. These problems are exemplary for dilemmas that autonomous
agents are faced with when interacting with humans. Furthermore, we argue that
a Newcomb-like formulation is more adequate in the human-machine interaction
case and demonstrate empirically that the unmodified Reinforcement Learning
algorithms end up with the well known maximum expected utility solution.
\end{abstract}}


\maketitle

\IEEEdisplaynontitleabstractindextext

%
\IEEEpeerreviewmaketitle

\IEEEraisesectionheading{\section{Motivation}\label{sec:introduction}}
Autonomous Unmanned Underwater Vehicles (UUVs) are used for a wide range of 
oceanographic, maritime mining, and military tasks including underwater surveys, 
inspection and maintenance of submerged structures, tracking oceanographic 
features, and undersea mapping to name a few (cf. \cite{seto2012marine}). 
Depending on their task, the physical shape of UUVs can be the traditional 
torpedo-shaped bodies or more like Remotely Operated Vehicles (ROVs) 
with many manipulators, cameras, and lights. 

The difficulty with autonomous submarines is that there is no communication 
link between a human operator due to the fact that radio waves cannot penetrate 
water very far. Therefore, the UUV loses its GPS signal as soon as it 
enters the water. Navigation is mostly performed using compasses, depth sensors, 
accelerometers, and sonars. The sensor information provides enough information 
for navigation using dead reckoning but are not sufficient to make an informed 
decision. 

In the upcoming deep sea mining scenarios, where wellheads are built
as subsea systems directly on the sea floor, ROVs and UUVs are used to 
build and maintain those structures. As direct human control is only 
possible with tethered ROVs, an increased usage of UUVs to maintain 
subsea systems is desirable. Using underwater acoustic positioning 
systems (long-baseline (LBL) systems), UUVs are able to find subsea 
structures like wellheads and processing systems. Only acoustic 
communication allows limited communication like broadcasting alarms. 

In such a scenario an unmanned underwater vehicle needs advanced 
autonomous decision making algorithms for task planning and behavior 
selection. Furthermore, in case of multiple cooperating robots and 
operator controlled vehicles, strategies for cooperating and 
solo actions have to be developed and risk analysis is vital 
\cite{thieme2015risk}. Besides traditional risk assessment using 
event and fault tree analysis, risk can also be assessed using examples
from game theory like the Prisoner's Dilemma or the Newcomb's Problem. 
We investigate these examples by applying Reinforcement Learning (RL).


Reinforcement Learning, one classic decision making and learning technique
in the field of autonomous learning, fits the setting of the two dilemmas very well,
and has been already applied in basic Prisoner's Dilemma scenarios. In
\cite{claus:drlcms}, the authors apply RL to multi-player domains, where cooperation
is beneficial and investigate the capability of two agents successfully establishing
a stable equilibrium strategy. The authors of \cite{bazz:lcipd} present the performance
of RL algorithms in the Prisoner's Dilemma, admitting knowledge of the problem 
structure. They distinguish between independent learners, where each agent has no
knowledge about the state of the other agents, and multi-agent settings, where
a joint decision is reached. They aim at establishing cooperation, which can be 
reached with a biased exploration strategy. Sandholm et al. \cite{sand:mrlipd} study 
the play of a RL agent against a fixed opponent strategy and itself. They manage to
achieve optimal play by using a history of moves and a representation of the move 
history by a neural network as the state.  On the contrary, Flache et al.
\cite{flache:scpl} take a different approach, which tries to explain the cooperative
behavior observed in experiments with a general psychologically inspired reinforcement
learning model.
\section{Modelling Deep Sea Robots using Newcomb-like Problems}
Newcomb's problems arise when an autonomous agent is in a situation where 
others have knowledge about its decision process via some mechanism 
(e.g. a statistical based model) that is not under its direct control. 
Newcomb-like problems cannot be handled by the conventional Causal 
Decision Theory, as the independence of the two decision makers is violated. 
Also most real decisions humans face are Newcomb-like, at least whenever other 
humans are involved. People automatically involve their experience and read 
unconscious or unintentional signals in order to build an internal social model 
of how someone decides. Simultaneously, they use those models to make their 
own choices and this when Causal Decision Theory fails. 

In general, real world decision scenarios can be often described as Newcomb-like 
problems in several ways. We however, do not assume a priori knowledge of the  
problem structure and instead treat the decisions made as a variation of a two-armed 
bandit. Furthermore, we restrict ourselves to the setting of independent learners. 
Namely, two actors, which can be characterized by their probability of cooperation 
in the respective scenario. This is possible, as we can establish the equivalence 
of this specific version of the Prisoner's Dilemma with Newcomb's Problem.
\subsection{Prisoner's Dilemma in Deep Sea Repair Robots}
In our example of a field of multiple oil wells and two available 
maintenance robots, two failures with smaller leakages occur at the same time. 
Each leakage can be fixed by one robot, but there is a chance of the robot to be
damaged in the process. Also, if the leakages are both not fixed, there will
be a cascade of events that will cause an even bigger oil spill. If both 
robots fix both leaks, then there will be no oil spill but still the chance
of slight damage to the robot remains. If one robot tries to fix one spill
and the other decides not to, then there will be no natural desaster, but
the second leak will still remain. Also the robot trying to fix the spill will
be severely damaged as the second spill causes complications with the first
leak. Underwater, the robots have no communication to coordinate whether they
will go for a repair or not, but can sense the location of the leaks and the
other robots with a sonar.

An exemplary payout matrix (regrets are treated as negative payouts) for 
this problem is depicted in Table~\ref{tab:oil-spill-symmetric-prisoners-dilemma}, 
with $R = -2.000$, the cost of the repair and eventual minor damages of the robot,
$S = -4.000$, the cost of major damages to the robot due to the overpressure
from the other leak as well as the costs for the oil spilled, $T = -1.000$,
the costs of the oil spill by not fixing the leak, $P = -3.000$, the cost of 
the major oil spill due to both leaks not getting fixed. Clearly, it satisfies
all the prereqisites to be a version of the Prisoner's Dilemma with having
the two conditions
\begin{equation}
T > R > P > S \qquad \text{and} \qquad R > \frac{T + S}{2}.
\end{equation}
fulfilled.
\begin{table}[bth]
    \begin{center}
    \caption{Oil Spill Prisoner's Dilemma Robot-Robot Regrets.}
    \label{tab:oil-spill-symmetric-prisoners-dilemma}
    \begin{tabular}{|c|c|c|c|}
        \cline{3-4}
        \multicolumn{2}{c}{} & \multicolumn{2}{|c|}{Robot 2} \\
        \cline{3-4}
        \multicolumn{2}{c|}{} & Repair & No Repair \\
        \cline{1-4}
        \multirow{2}{*}{Robot 1} & Repair & (R, R) & (S, T) \\
        \cline{2-4}
        & No  Repair & (T, S) & (P, P) \\
        \hline
    \end{tabular}
    \end{center}
\end{table}
The dilemma unfolds as follows: since the temptation $T$ is the lowest regret
a robot can receive while the sucker's regret $S$ is the highest, the
optimal strategy for a robot locally would be to always not repair (defect) since
the action of the other robot is unknown. The robot therefore has
no incentive to change its decision, if it is unclear what the other
will do. This situation is also known as Nash equilibrium in game theory.

If both robots would know what the other does or would be able to
communicate about a common strategy, then clearly mutual cooperation (both repair) would
be the optimal thing to do. Herein lies the dilemma. Locally, it is 
optimal for each robot to not repair while globally it would be optimal
to repair each leak. For this dilemma to occur, we need the first condition
to be fulfilled. Then the second condition prevents taking turns at defection
and cooperation to be more profitable in an iterated setting.

A robot could now concretely decide what to do using the expected utility
of each outcome. This means that it assigns a certain cooperation probability
$p$ to the problem, which gives the likelihood that the other robot will
choose \texttt{repair}. As the problem is symmetrical, we can write down the 
expected utility $EU$ as
\begin{equation}
\left\{
\begin{split}
    EU(\text{repair}) &= p * \text{$(-2.000)$} + (1-p) * \text{$(-4.000)$}, \\
    EU(\text{no repair}) &= (1-p) * \text{$(-1.000)$} + p * \text{$(-3.000)$}.
\end{split}\right.
\end{equation}
If therefore the chance of the other robot cooperating is greater than 
$p = 0.75$ then the robots will cooperate and both choose \texttt{repair}.
\subsection{Imbalance in Prisoner's Dilemma due to Human Interaction}
If now a human happens to be involved in the dilemma, the situation changes
drastically. Suppose a maintenance worker is underway in a single person
submarine to carry out maintenance tasks that cannot be handled by 
autonomous robots yet. In this moment the situation described prior unfolds
and two leaks have to be fixed. Compared to the regret, when a robot gets damaged,
the regret, when the submarine carrying the human gets damaged,
is much higher. The event in which the human solely decides to fix one leak
and an accident happens is now valued with a regret of $-1.000.000$. The 
modified payout matrix is written down in 
Table~\ref{tab:oil-spill-asymmetric-prisoners-dilemma}.
\begin{table}[htb]
    \begin{center}
        \caption{Oil Spill Prisoner's Dilemma Robot-Human Regrets.}
        \label{tab:oil-spill-asymmetric-prisoners-dilemma}
        \begin{tabular}{|c|c|c|c|}
            \cline{3-4}
            \multicolumn{2}{c}{} & \multicolumn{2}{|c|}{Human} \\
            \cline{3-4}
            \multicolumn{2}{c|}{} & Repair & No Repair \\
            \cline{1-4}
            \multirow{2}{*}{Robot 1} & Repair & ($-2.000, -2.000$) & 
                ($-4.000, -1.000$) \\
            \cline{2-4}
            & No  Repair & ($-1.000, -1.000.000$) & ($-3.000, -3.000$) \\
            \hline
        \end{tabular}
    \end{center}
\end{table}
As can easily be seen, the regret situation is now highly skewed, since the
life of a human naturally is weighted much higher than the integrity of the 
replaceable robot. Therefore, the robot would still decide to repair according to 
expected utility starting
from a confidence of $p_r = 0.75$ that the human will also repair, whereas the
human would only decide to go down for a repair if she/he is almost sure with a
confidence of $p_h = 0.999$ that the robot will follow and do it's job.

\subsection{Modelling a Skewed Prisoner's Dilemma with Newcomb's Problem}

The classical Prisoner's Dilemma cannot fully accompany skewed problems of this
type. But fortunately, as Lewis stated \cite{lewi:pdinc}, the Prisoner's
Dilemma can be seen as two coupled Newcomb's Problems. In the original description
of the Newcomb's Problem, a superhuman intelligence is predicting the choice
the player is going to make and placing the bets on the two options available
accordingly. The coupled Newcomb's Problem from the viewpoint of the Robot in
the asymmetrical dilemma can be observed in 
Table~\ref{tab:newcomb-payouts-oil-spill-robot-view}.
\begin{table}[htb]
    \begin{center}
        \caption{Oil Spill Dilemma Newcomb Version, Robot Viewpoint.}
        \label{tab:newcomb-payouts-oil-spill-robot-view}
        \begin{tabular}{|c|c|c|c|}
            \cline{2-3}
            \multicolumn{1}{c}{} & \multicolumn{2}{|c|}{Robot} \\
            \cline{2-3}
            \multicolumn{1}{c|}{} & Repair & No Repair \\
            \cline{1-3}
            Predict: Human Repair & $-2.000$ & $-1.000$ \\
            \cline{1-3}
            Predict: Human No Repair & $-4.000$ & $-3.000$ \\
            \hline
        \end{tabular}
    \end{center}
\end{table}
In our version, instead of the superhuman intelligence predicting our choices,
we predict how the opponent (in this case the human) in the Prisoner's Dilemma
version of the problem will behave. Depending on the prediction accuracy, the
payout can be observed upon choosing one of the actions. Originally, the 
prediction probability was set very close to $p = 1.0$ and therefore the 
dilemma again manifests itself in that the expected utility recommends to
do the opposite of the dominating choice. By following the argumentation in
the work of Nozick \cite{nozi:nppc}, the robot now can choose to not repair,
since both outcomes of not repairing are better than their alternatives, no
matter if the prediction was correct or not. If we insert the numbers in the
above formulas for expected utility then, on the other hand, with a very high
probability that the prediction was correct, it is better to choose to repair.

Using this modified problem formulation, we can now write down the regrets
from the viewpoint of the human in 
Table~\ref{tab:newcomb-payouts-oil-spill-human-view}. 
\begin{table}[htb]
    \begin{center}
        \caption{Oil Spill Dilemma Newcomb Version, Human Viewpoint.}
        \label{tab:newcomb-payouts-oil-spill-human-view}
        \begin{tabular}{|c|c|c|c|}
            \cline{2-3}
            \multicolumn{1}{c}{} & \multicolumn{2}{|c|}{Human} \\
            \cline{2-3}
            \multicolumn{1}{c|}{} & Repair & No Repair \\
            \cline{1-3}
            Predict: Robot Repair & $-2.000$ & $-1.000$ \\
            \cline{1-3}
            Predict: Robot No Repair & $-1.000.000$ & $-3.000$ \\
            \hline
        \end{tabular}
    \end{center}
\end{table}
As these are now two decoupled problems, it is additionally possible to assign
different cooperation probabilities to the two involved entities. Usually,
we would assign a very high cooperation probability to a robot as it should do
what it is supposed to do, but in the setting of autonomous decision making
with limited communication under water, we cannot be so sure anymore. 
Also for humans, depending on the situation, cooperation probability can 
exhibit a great variance. 
\begin{figure*}[t]
    \begin{center}
        \begin{subfigure}[t]{0.3\textwidth}
            \includegraphics[width=\textwidth]{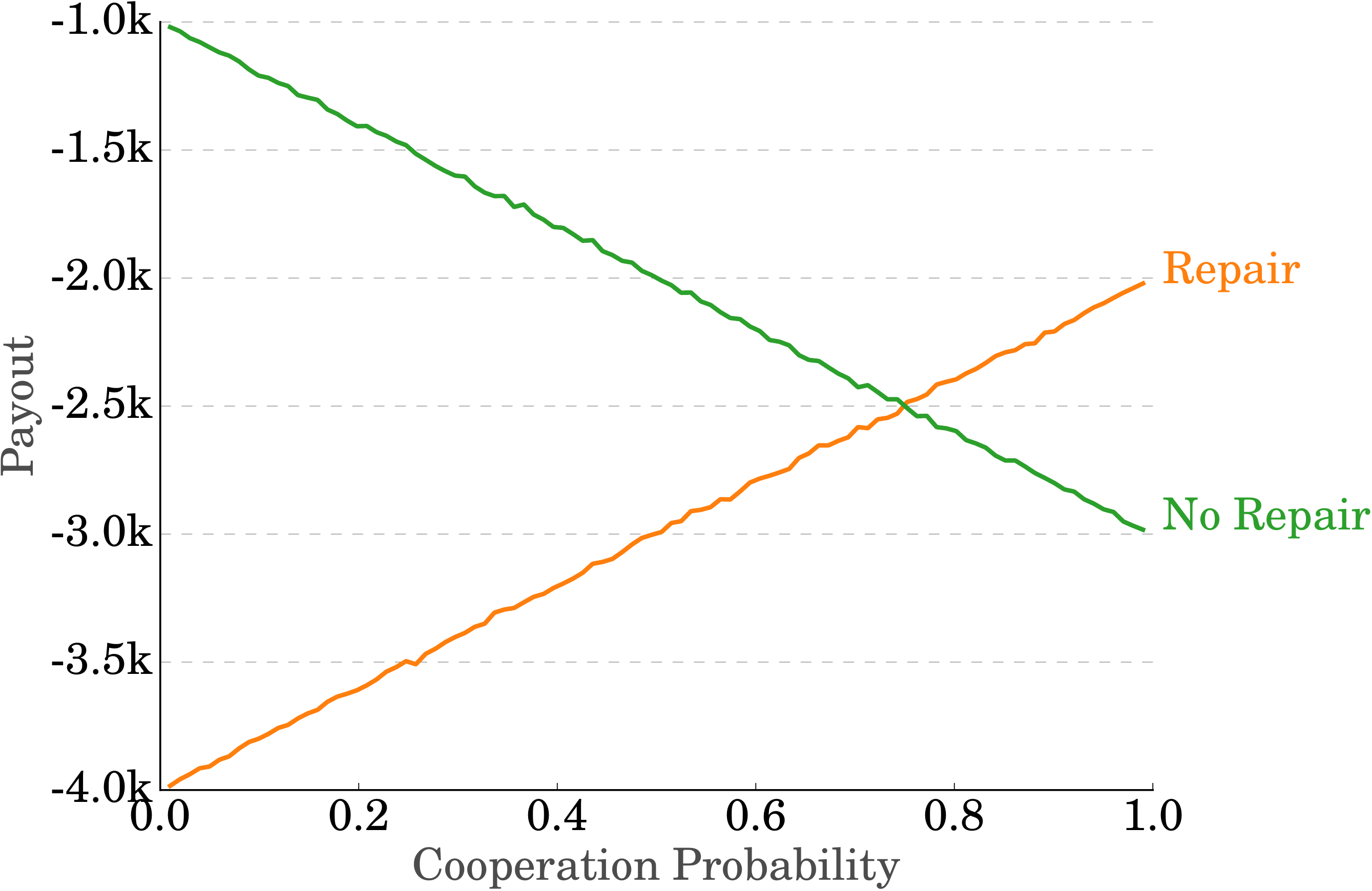}
            \caption{Payout for two baseline agents that always repair or 
                don't repair.}
            \label{fig:rlcomb-nc-a}
        \end{subfigure}
        \hspace{1mm}
        \begin{subfigure}[t]{0.3\textwidth}
            \includegraphics[width=\textwidth]{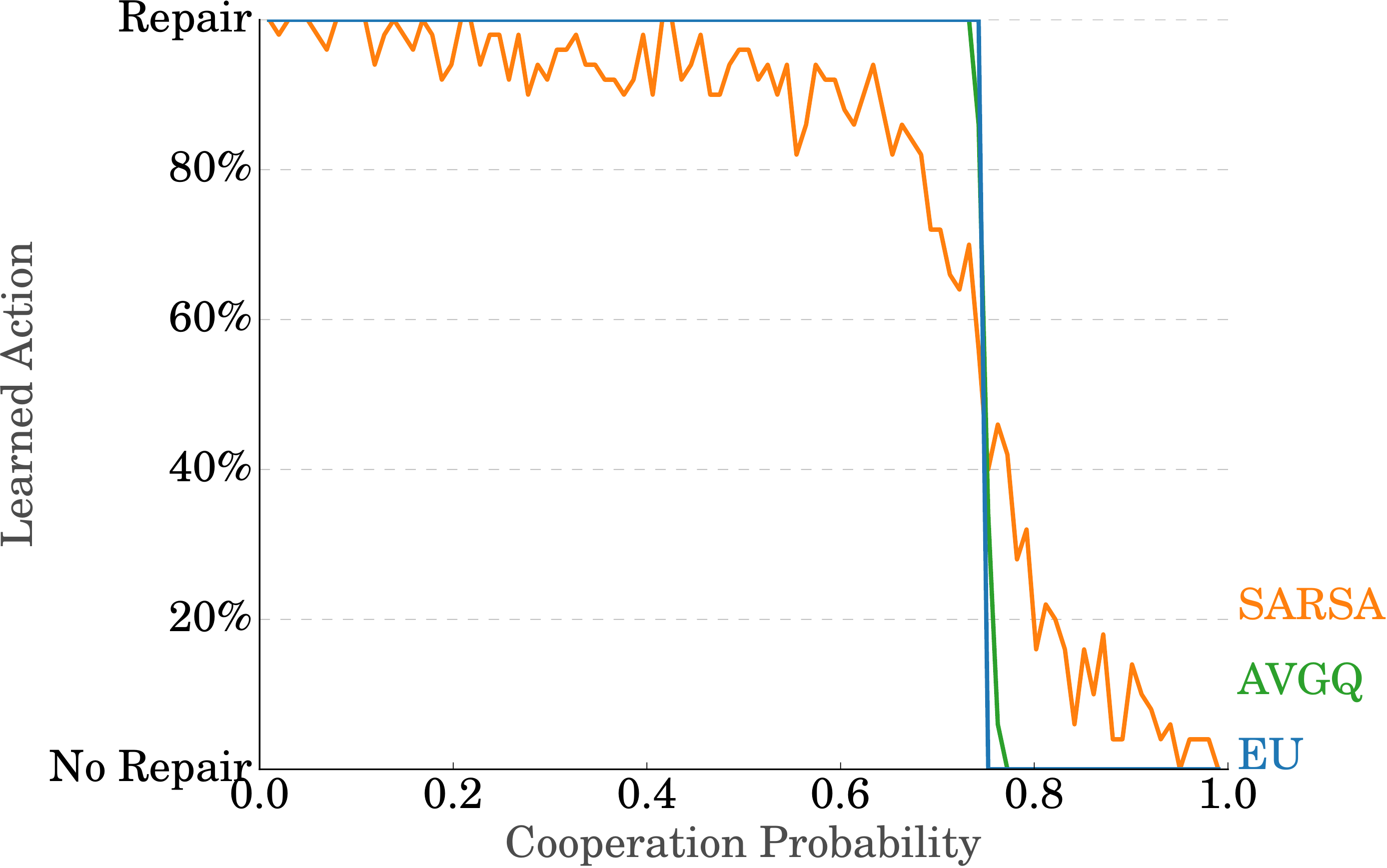}
            \caption{Learned action for SARSA with $\alpha = 0.1, \gamma = 0.9, \epsilon = 0.1$, AVGQ and EU. Probability of repairing on y-axis.}
            \label{fig:rlcomb-nc-b}
        \end{subfigure}
        \hspace{1mm}
        \begin{subfigure}[t]{0.3\textwidth}
            \includegraphics[width=\textwidth]{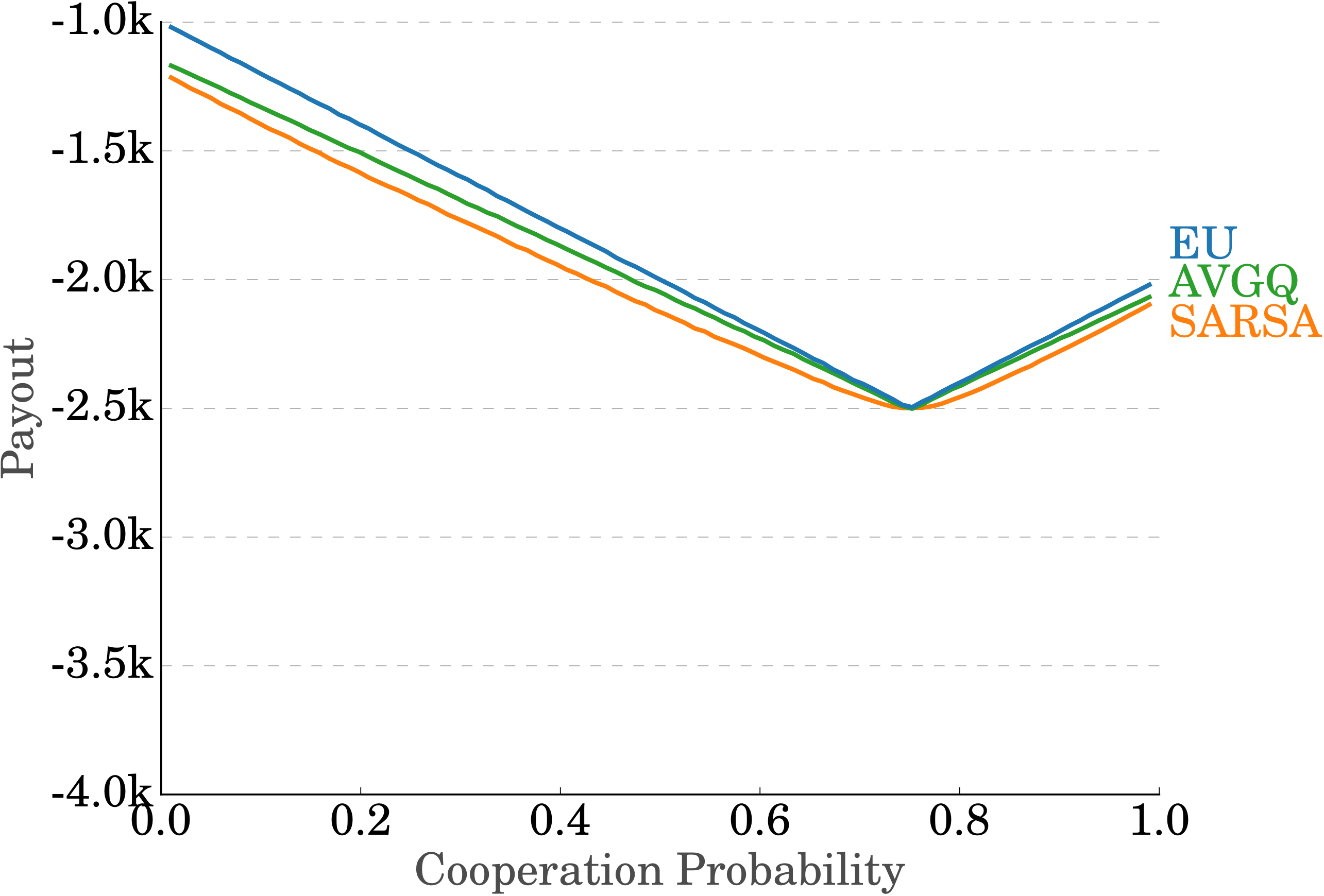}
            \caption{Payout for SARSA (same parameters), AVGQ and EU for varying cooperation probability.}
            \label{fig:rlcomb-nc-c}
        \end{subfigure}
    \end{center}
    \vspace{-3mm}
    \caption{Experimental results for Newcomb's Problem, Robot View.}
\end{figure*}
\section{Simulation Results using Reinforcement Learning}
To assess how an autonomous robot would learn to decide in a deep sea repair
scenario, we can now use the above models to simulate behavior. The 
autonomous learning technique we chose to test is unmodified Reinforcement 
Learning, to get a baseline of the most popular autonomous learning
algorithms that currently exist in the literature.

It is important to notice that in our setting of decision making, one 
decision maker plays against her/his opponent with a fixed cooperation or 
prediction probability. In other words, there is no state information required 
for both Newcomb's or the Prisoner's Dilemma type of formulation. Hence, we
model our problems as playing a bandit. Specifically, for each action $a \in
A_{n} := \{\texttt{repair}, \texttt{ no repair}\}$ we maintain the $Q$ function
without the state variable as $Q(a) = \mathbb{E}\left[\sum_{t} \gamma^t
r_t\right]$, which represents the future expected discounted reward, where 
$t$ is the current iteration step, $\gamma \in [0,1]$ a discounting factor and
$r_t$ the payout received at iteration step $t$. The two problems are evaluated
in an iterated fashion. This means that the agent is not faced with the
decision only once, but for $N$ times and can therefore learn from those
interactions.

For updating the $Q$ values, we use the well known SARSA \cite{sutton:rl}
algorithm with learning rate $\alpha$. A second learning agent is calculating
averaged payouts for the $Q$ function (AVGQ). A third agent calculates expected
utility (EU) for comparison. As the action selection mechanism, we chose an
$\epsilon$-greedy strategy, where the action with the largest corresponding 
$Q$ value is chosen most of the times and a random action with probability
$\epsilon$. In each problem, learning was studied for $N = 10000$ iteration
steps and averaged over 50 independent runs.

For the Newcomb formulation, we additionally implemented two baseline agents,
that either always repair or don't repair. If now the prediction accuracy is
varied, the payout behaves as expected, which can be seen in
Figure~\ref{fig:rlcomb-nc-a}.

For the SARSA agent, we can observe adaptation to the problem depending on 
the prediction accuracy. In Figure~\ref{fig:rlcomb-nc-b}, we can observe that 
the agent will always choose \texttt{no repair} if the cooperation probability is close
to zero, and always \texttt{repair} if the probability is very high. If this probability
accuracy is approximately $0.75$, then the agent cannot learn what to do and
chooses actions at random. The RL agent therefore is able to learn the correct 
behavior depending on the environmental parameters. The average $Q$ (AVGQ)
implementation learns to behave in the same way, even closer to the expected
utility solution. In accordance to this behavior, the payout varies as shown in
Figure~\ref{fig:rlcomb-nc-c}.

What we can conclude from Figures~\ref{fig:rlcomb-nc-b} and 
\ref{fig:rlcomb-nc-c} is that the RL agent is able
to learn the action that maximizes the payout and corresponds with the expected
utility solution. This means in a consistent environment where the partner almost always
cooperates or does the opposite, learning will succeed. For cooperation probabilities
around the expected utility threshold ($p = 0.75$), the behavior is not perceived
as consistent and the best the SARSA agent can do is to choose actions in a random
manner.

\begin{figure*}[t!]
    \begin{center}
    \begin{subfigure}[t]{0.3\textwidth}
        \includegraphics[width=\textwidth]{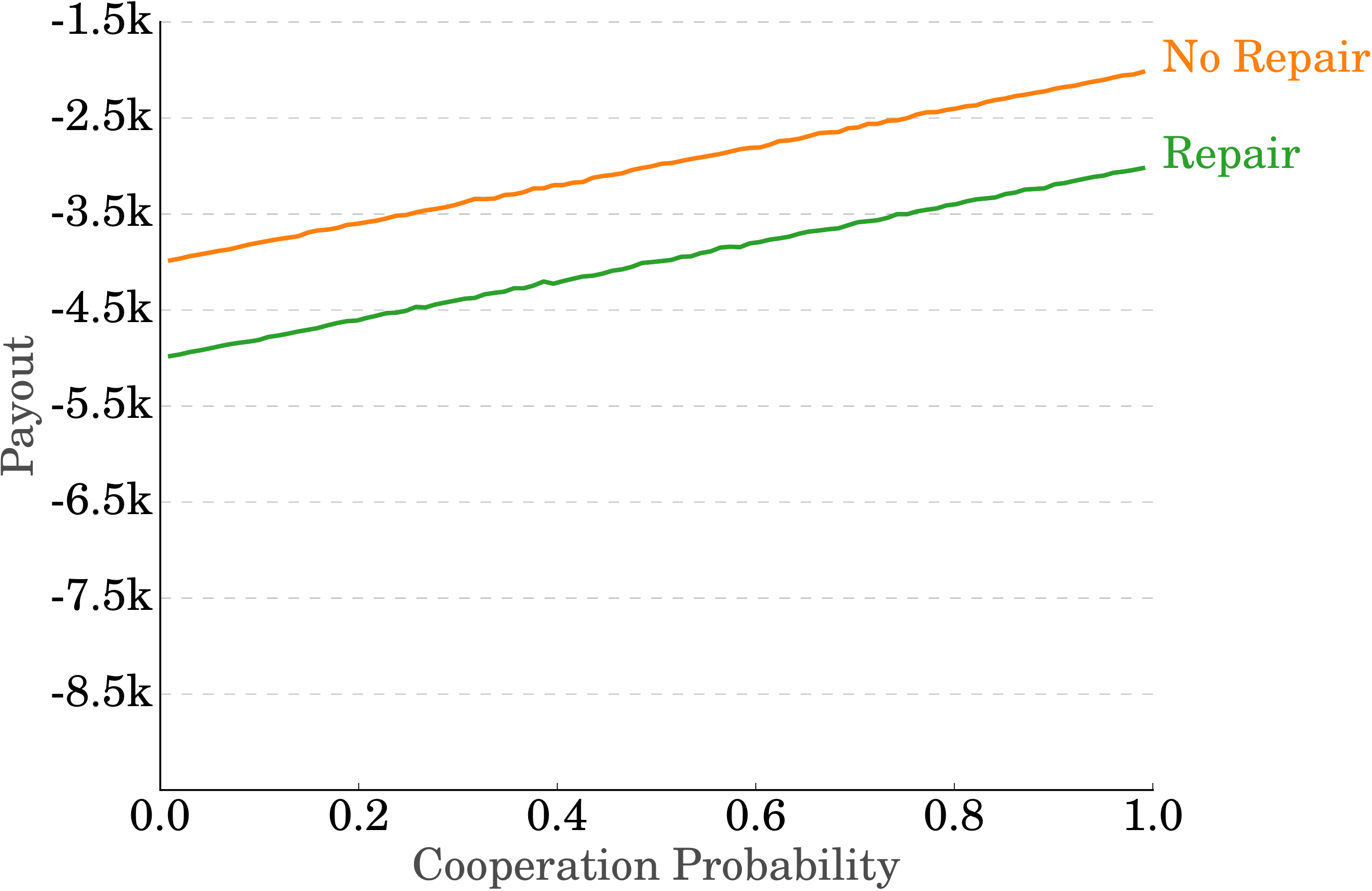}
        \caption{Payout for two baseline agents that always repair/don't 
                repair.}
        \label{fig:rlcomb-pd-a}
    \end{subfigure}
    ~
    \begin{subfigure}[t]{0.3\textwidth}
        \includegraphics[width=\textwidth]{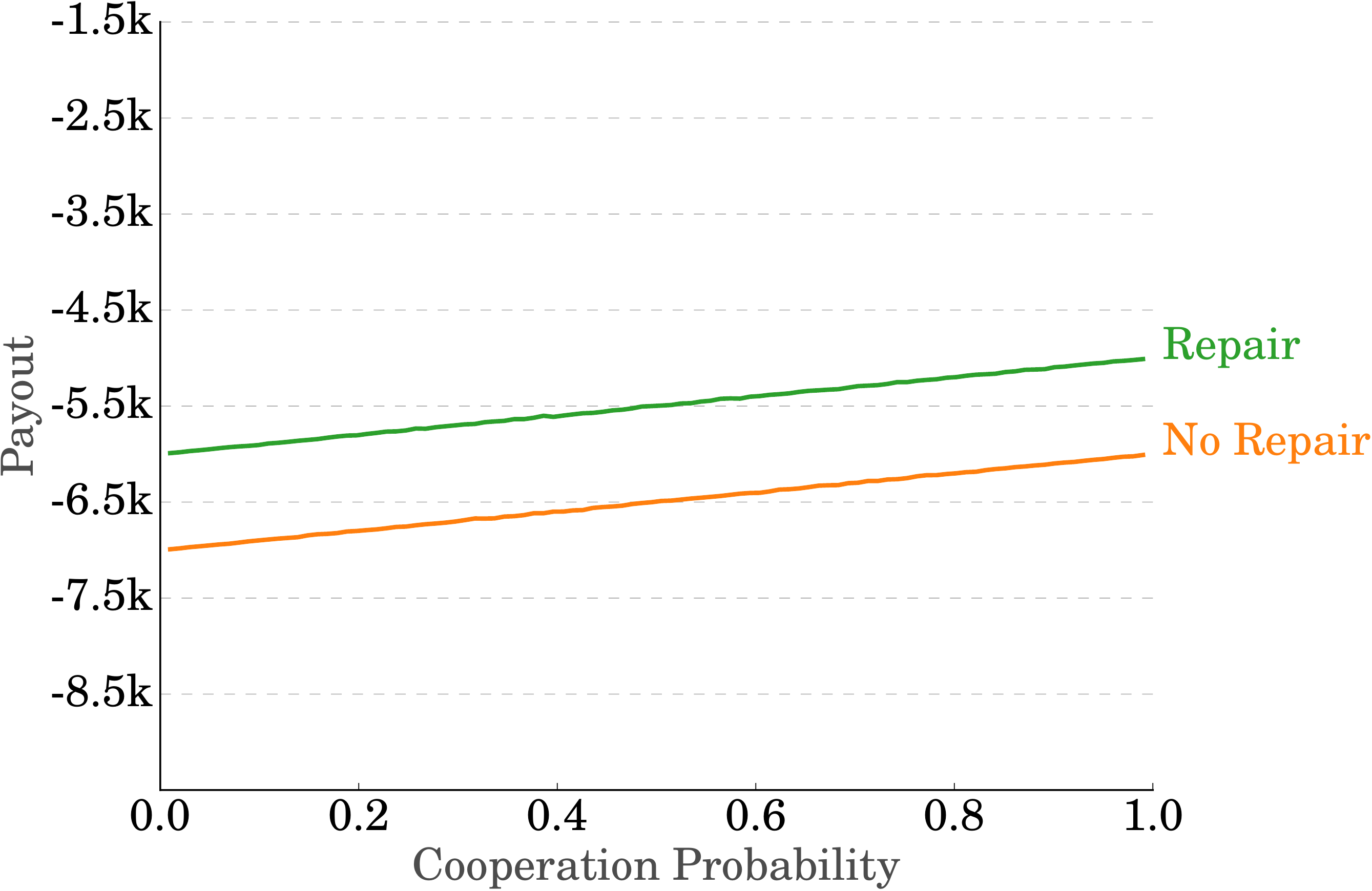}
        \caption{Sum of payouts for the baseline agents.}
        \label{fig:rlcomb-pd-b}
    \end{subfigure}
    ~
    \begin{subfigure}[t]{0.3\textwidth}
        \includegraphics[width=\textwidth]{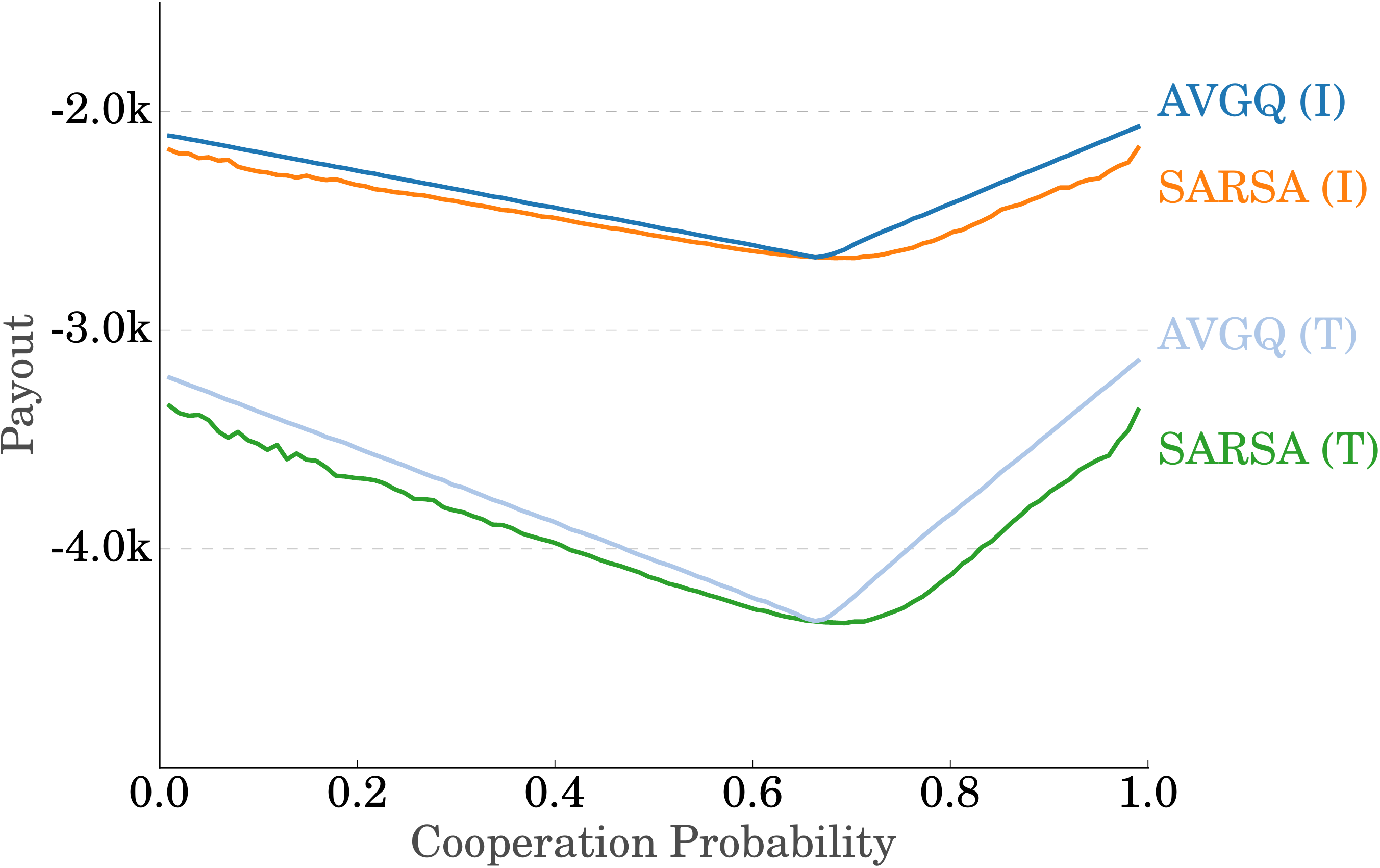}
        \caption{Payout for SARSA with $\alpha = 0.1, \gamma = 0.9,
            \epsilon = 0.1$ and AVGQ. (I) individual, (T) sum of payouts.}
        \label{fig:rlcomb-pd-c}
    \end{subfigure}
    \end{center}
    \vspace{-3mm}
    \caption{Experimental results for the Prisoner's Dilemma formulation.}
\end{figure*}

In the Prisoner's Dilemma formulation, we can imagine two modes of operation.
Namely, either each agent receives only its own regret, or both agents receive
the sum of the two regrets. In Figures~\ref{fig:rlcomb-pd-a} and 
\ref{fig:rlcomb-pd-b}, it can be seen that for individual payouts \texttt{no repair}
always dominates \texttt{repair}, while for the sum of payouts \texttt{repair} dominates.
This reflects the dilemma in the underlying problem and it would therefore
be optimal for the RL agents to learn either to not repair (in the individual
payout setting) or to repair (in the sum-of-regret setting). This is in fact
the case if we consider the results from Figure~\ref{fig:rlcomb-pd-c}, in which
the regrets for the SARSA and AVGQ agents are depicted for the individual (I)
and sum (T) regret setting.
%
%
\section{Conclusion}
In this abstract, we study the behavior of basic unmodified reinforcement learning
agents, when faced with decision theoretic thought experiments. In both problems,
Newcomb's Problem and Prisoner's Dilemma, RL learning algorithms learned to take
actions according to the maximum expected utility solution. This is due to the fact
that RL maximizes the cumulative expected reward, which is in these settings
similar to the expected utility. In some situations, it might be not desirable to
decide according to utility, therefore other techniques from causal decision theory
are to be investigated in conjunction with learning algorithms from the field
of autonomous systems.

Most existing works in the literature attempt to steer RL agents towards favourable
decision equilibria by the use of modified RL algorithms. Our present results have
verified that both rewarding procedure (what to reward) and the reward structure 
(how do we reward) are the most crucial points in shaping the agents decision. It is 
then beneficial to investigate further into reward shaping mechanisms or other means 
of rewarding. One example could be the integration of moral values, as proposed in
\cite{edal:dgpd}. The research in this direction could open the field of RL to a much
broader philosophical discussion in decision making.

In respect of robotics, our results show how autonomous action planning in conflicting 
situations can be simulated. Adjusting the cooperation probabilities according to 
statistical data enables the simulation of different assumptions about a potential 
cooperation partner. The technique can be integrated into future action planning 
algorithms of service and maintenance robots. It is also conceivable to use it in 
autonomous cars to assess different traffic situations. The car would be able to 
simulate different outcomes of maneuvers according to statistical data for the 
probabilities of breaking or obeying traffic rules.  
\end{document}